\documentclass{article}

\usepackage{arxiv}

\usepackage[utf8]{inputenc}
\usepackage[T1]{fontenc}
\usepackage[colorlinks=true,allcolors=black]{hyperref}
\usepackage{url}
\usepackage{booktabs}
\usepackage{amsfonts}
\usepackage{amsmath}
\usepackage{nicefrac}
\usepackage{microtype}
\usepackage{graphicx}
\usepackage[numbers]{natbib}
\usepackage{doi}
\usepackage{float}
\usepackage{xcolor}
\usepackage{tikz}
\usepackage{caption}
\usepackage{tabularx}
\usepackage{comment}
\usetikzlibrary{backgrounds}
\usepackage[table]{xcolor}

\usepackage{tcolorbox}
\tcbuselibrary{skins}

\title{HiPO: Hierarchical Preference Optimization for Adaptive Reasoning in Large Language Models}

\author{
  Darsh Kachroo \\
  Vellore Institute of Technology, Chennai \\
  \texttt{darsh.kachroo2022@vitstudent.ac.in} \\
  \And
  Adriana Caraeni \\
  University of Massachusetts Amherst \\
  \texttt{acaraeni@umass.edu} \\
  \And
  Arjun Prasaath Anbazhagan \\
  Northwestern University \\
  \texttt{arjunanbazhagan2026@u.northwestern.edu} \\
  \AND
  Brennan Lagasse \\
  Yale University / Algoverse AI Research \\
  \texttt{brennan@algoverseairesearch.org} \\
  \And
  Kevin Zhu \\
  Algoverse AI Research \\
  \texttt{kevin@algoverse.us}
}

\date{}

\renewcommand{\headeright}{}
\renewcommand{\undertitle}{}
\renewcommand{\shorttitle}{HiPO: Hierarchical Preference Optimization for Adaptive Reasoning in LLMs}

\begin{document}
\maketitle

\begin{abstract}
Direct Preference Optimization (DPO) is an effective framework for aligning large language models with human preferences, but it struggles with complex reasoning tasks. DPO optimizes for the likelihood of generating preferred over dispreferred responses in their entirety and lacks the granularity to provide feedback on subsections of many-step solutions typical of reasoning tasks. Existing methods excel at either stable preference learning (e.g., DPO variants like KTO and RSO) or structured reasoning (e.g., ReMA's multi-agent RL framework, Tree of Thoughts), but fail to merge these complementary strengths. We propose HiPO (Hierarchical Preference Optimization), an extension of DPO that separates responses into reasoning segments (query clarification and context, reasoning steps, and answer) and computes loss as a weighted sum of the DPO loss for each segment. Our approach enables segment-specific training while maintaining DPO's computational efficiency and training stability. We demonstrate that for multiple 7B LLMs fine-tuned using HiPO and DPO on the Math Stack Exchange preference dataset, the models trained with HiPO outperform the others on a variety of common math benchmarks and achieve greater organization, logical flow, and consistency as measured by GPT-4.1.
\end{abstract}


\section{Introduction}

\subsection{Motivation}
Direct Preference Optimization (DPO) efficiently aligns large language models with human preferences by directly optimizing pairwise preference data, reducing required computation costs. This method is an alternative to reinforcement learning approaches, which are extremely resource-intensive. However, DPO's monolithic treatment of responses---lacking component-level granularity---creates a fundamental limitation: it cannot separately optimize different reasoning components (query clarification, structured reasoning, and answer formulation), preventing practitioners from targeting specific deficiencies or adapting training emphasis to problem-specific reasoning demands.

Different reasoning tasks naturally require different cognitive skills. A problem with an unclear or ambiguous question demands careful query interpretation prior to any solution attempt. Complex, multi-step problems requiring significant domain knowledge benefit from structured meta-cognitive reasoning: the ability to plan, organize steps, and apply relevant concepts systematically. Additionally, tasks requiring specific output formats, such as competition submissions or standardized test responses, require precise final answer formulation. Standard DPO optimizes all aspects as one, and does not natively support targeted improvement of specific reasoning deficiencies.

HiPO addresses this limitation by decomposing responses into distinct reasoning segments and enabling practitioners to control training emphasis through adjustable segment weights. This allows prioritizing query understanding for problems with ambiguous specifications, strengthening step-by-step reasoning for complex analytical tasks, or building complex reasoning capabilities incrementally by first developing query interpretation, then meta-cognitive planning, and finally integrating both. This controllability allows models to develop distinct reasoning competencies in a targeted manner, with different model architectures potentially responding better to different training strategies and providing adjustable mechanisms for improving reasoning capabilities where they are most needed.

HiPO maintains DPO's computational efficiency as a single-agent, single-pass framework while enabling this fine-grained optimization through decomposition of the preference score into segment-level components.

\subsection{Applications}
Models with stronger reasoning capabilities have wide-ranging applications. In particular, tasks that require detailed analysis, multi-step planning, and reduced susceptibility to hallucinations stand to benefit the most. Such models would not only generate correct answers, but also provide transparent explanations of their reasoning process. This makes them especially suited for higher-level, abstract queries---for example, instructions of the form ``Design a system that accomplishes X,'' where the model must decompose the request into subproblems and outline a coherent plan without extensive prompting.

Beyond general reasoning improvements, the hierarchical decomposition introduced in HiPO also holds promise for information retrieval and retrieval-augmented generation systems. Unlike purely generative alignment tasks, retrieval-oriented reasoning involves evidence selection and synthesis, where models must interpret complex queries and integrate information dispersed across large input contexts. These abilities are especially advantageous for retrieval scenarios, where understanding the query, reasoning over extensive retrieved content, and producing relevant, grounded outputs are distinct yet interdependent stages that together determine retrieval quality.

\subsection{Contributions}
\begin{enumerate}
    \item \textbf{Hierarchical Response Decomposition}: We propose a procedure for augmenting preference datasets by decomposing each response into three parts: (1) restatement of the question and context, (2) reasoning towards the answer, and (3) the final answer. We believe this enables targeted optimization of distinct reasoning skills without requiring multi-pass inference or complex multi-agent architectures.

    \item \textbf{Multi-objective DPO Extension}: We extend Direct Preference Optimization with segment-level auxiliary losses, creating a weighted objective that optimizes preferences across both individual reasoning components and the complete response.
\end{enumerate}

\section{Related Work}

\subsection{The Evolution from RLHF to DPO}

The alignment of large language models with human preferences has undergone significant evolution, moving from complex reinforcement learning frameworks toward more direct optimization approaches. Traditional Reinforcement Learning from Human Feedback (RLHF) \cite{christiano_deep_2023,ouyang_training_2022} requires training separate reward models and employing unstable policy optimization algorithms like PPO. As a result, fine-tuning with RLHF is unstable and has high computational costs.

Direct Preference Optimization (DPO) \cite{rafailov_direct_2024} eliminated the need for explicit reward modeling, instead directly optimizing policy models on preference data through a theoretically grounded classification objective. However, DPO treats responses as monolithic units, applying uniform preference learning across entire outputs without considering the hierarchical nature of complex reasoning tasks.

Recent DPO variants have attempted to address various limitations while maintaining the direct optimization paradigm. Statistical Rejection Sampling Optimization (RSO) \cite{liu_statistical_2024} improves upon DPO by better approximating the optimal policy through rejection sampling, while Kahneman-Tversky Optimization (KTO) \cite{ethayarajh_kto_2024} incorporates human cognitive biases into the preference learning objective. Despite these advances, all existing DPO variants share a fundamental limitation: they optimize preferences at the response level without decomposing the underlying reasoning process into learnable components. This limitation prevents the targeted optimization of specific reasoning deficiencies, such as unclear problem queries or difficult reasoning tasks.

\subsection{The Quest for Better Reasoning: From Prompting to Meta-Cognition}
Parallel to advances in preference optimization, significant progress has been made in eliciting sophisticated reasoning from language models through structured prompting and meta-cognitive frameworks. Chain-of-Thought prompting \cite{wei_chain--thought_2023} demonstrated that explicit step-by-step reasoning dramatically improves performance on complex tasks, while Tree of Thoughts \cite{yao_tree_2023} extended this concept by exploring multiple reasoning paths through deliberate search. Self-Consistency \cite{wang_self-consistency_2023} further showed the benefits of sampling diverse reasoning trajectories and aggregating results.

However, these prompting-based approaches rely on fixed templates and predefined reasoning structures, limiting their adaptability across diverse problem domains. Recent work has moved toward more flexible meta-cognitive frameworks that enable models to reason about their own reasoning processes. Self-Refine \cite{madaan_self-refine_2023} enabled iterative improvement through self-feedback loops. Shinn et al.\ \cite{shinn_reflexion_2023} demonstrated how models can learn from their mistakes through structured self-reflection.

The most sophisticated approaches have employed multi-agent reinforcement learning to achieve hierarchical meta-thinking. ReMA \cite{wan_rema_2025-1} proposed a framework with high-level planning agents and low-level execution agents, showing promising results in complex reasoning tasks. However, these RL-based meta-cognitive approaches suffer from the same training instabilities and computational overhead that originally motivated the development of DPO as an alternative to RLHF.

\subsection{The Preference Learning Gap in Hierarchical Reasoning}
The evolution of both preference optimization and reasoning frameworks reveals a critical gap in current approaches. While DPO and its variants have demonstrated superior training stability and efficiency compared to RLHF, they remain fundamentally limited by their monolithic treatment of responses. Conversely, hierarchical reasoning approaches like ReMA \cite{wan_rema_2025-1} and multi-agent frameworks \cite{bilal_meta-thinking_2025} recognize the importance of decomposing complex reasoning into structured components, but rely on reinforcement learning objectives that reintroduce the very training instabilities that DPO was designed to eliminate.

This fundamental tension creates a significant limitation in current preference learning approaches: existing methods either apply stable preference optimization to flat response structures, or employ hierarchical reasoning with unstable training paradigms. Recent work on hierarchical preference optimization \cite{singh_direct_2025} has begun to explore sub-goal decomposition, but focuses primarily on task-level hierarchies rather than reasoning process decomposition.

The absence of a framework that combines the training stability of direct preference optimization with the structured reasoning decomposition of meta-cognitive approaches represents a significant gap in the literature. Current methods fail to leverage human preferences about reasoning processes themselves---preferences about how models should plan, reflect, and structure their thinking---rather than just preferences about final outputs. This limitation is particularly acute for complex reasoning tasks where the quality of the thinking process is as important as the correctness of the final answer.

\section{Methods}

\subsection{Data Generation}
We use a preference dataset derived from questions and answers on Stack Exchange \cite{preference_data_math_stack_exchange}. Questions on this site are often technical and require detailed, multi-step reasoning. As a result, we have reason to suspect the responses are sufficiently long and complicated to warrant breaking them down into separate segments. For each response, we prompt GPT-4.1 to generate each of the following segments consistent with the original response; details regarding the specific prompt used can be found in Appendix~\ref{sec:GPT_dataset_prompt_inst}:

\begin{enumerate}
    \item $R_q$ (Refined Query) ($y_{R_q}$) --- A restatement of the query $q$ that contextualizes the response.
    \item $M_t$ (Meta-thinking) ($y_{M_t}$) --- A step-by-step reasoning or structured reasoning steps employed in the response.
    \item $A$ (Answer) ($y_{A}$) --- The final answer generated by the original response.
\end{enumerate}

These three segments are selected because they form a generalized, clear, and sequential reasoning structure that can map onto a wide variety of responses. Moreover, models optimized to produce these three reasoning segments can be used in sequence to generate a response that follows a clear reasoning structure.

For the sake of leveraging these segments for preference optimization, we assume that the responses in the dataset can be expressed in the given form without much distortion. However, since the segments are defined sufficiently broadly, we believe it should encapsulate nearly all reasoning traces. Under this assumption, we expect the reasoning segments associated with each preferred answer to be preferred over their counterparts associated with the corresponding dispreferred answer.

\begin{figure}[h!]
    \centering
    \includegraphics[width=0.2\columnwidth]{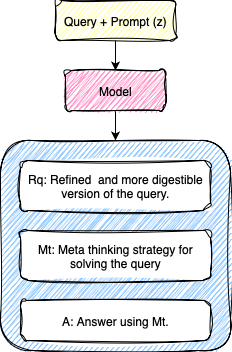}
    \caption{HiPO dataset creation: GPT-4.1 augments each preference pair with three structured segments.}
    \label{fig:dataset_creation}
\end{figure}

Unlike prior works that decouple query reformulation and reasoning into separate model passes \cite{wan_rema_2025-1}, we adopt a single-shot approach depicted in Figure~\ref{fig:dataset_creation}, where refined query ($R_q$), meta-thinking ($M_t$), and answer ($A$) are jointly generated in one trajectory. This couples the three segments and ideally ensures they are more consistent with one another.

\subsection{Segment-Specific Optimization}
We can separately optimize for the likelihood of the generation of any one of the three segments of a response using a DPO-like loss. We denote these different training settings as follows: (1) $R_q$-only optimization maximizes likelihood of the preferred refined query, (2) $M_t$-only focuses on meta-thinking steps, and (3) $A$-only optimization targets final answer generation. To formulate the loss for this, we need a few additional definitions.

\textbf{Log-Probability Computation:}
With an initial instruction prompt $p$ and query $q$, we let $z = \text{concat}(p, q)$ and generate the response $y$. Let $y = [y_1, \dots, y_n]$ and $\ell_i = -\log \pi_\theta(y_i \mid z, y_{<i})$. Also let $\pi_\theta$ be an LLM parametrized by $\theta$. Then
\[
\log \pi_\theta(y \mid z) = -\sum_{i=1}^T \ell_i
\]

\textbf{Segment-Level Log-Probability:} For each component $k \in \{R_q, M_t, A\}$, we define each segment as a tuple of start and end token indices $(s_k, e_k)$. Then
\[
\log \pi_\theta(y_k \mid z) = -\sum_{i=s_k}^{e_k} \ell_i
\]

\textbf{Segment-Level Difference in Log-Likelihood:}
For query $q$ with preferred answer $y^+$ and dispreferred answer $y^-$, the differences in the difference of log likelihoods under the reference and current models for segment $k$ is
\begin{align*}
\Delta_{k,z, y^+,y^-}(\theta) &= [\log\pi_\theta(y^+_k|z) - \log\pi_\theta(y^-_k|z) \\
&\quad - \log\pi_\text{ref}(y^+_k|z) + \log\pi_\text{ref}(y^-_k|z)]
\end{align*}

\textbf{Segment-Level Loss:} The loss for the generation of the segment alone, formulated as in \cite{rafailov_direct_2024}, is
\[
\mathcal{L}_k(\theta) = -\mathbb{E}_{(z,y^+,y^-)}[\log\sigma(\beta\Delta_{k,z,y^+,y^-}(\theta))]
\]

\textbf{Hierarchical Preference Optimization Loss:}
Given the segment-level losses, we construct the overall loss function as a weighted sum of the segment-level loss functions:
\begin{align}
\label{eq:main_loss}
\mathcal{L}(\theta) = \sum_{k \in \{R_q, M_t, A, y\}} w_k \cdot \mathcal{L}_k(\theta)
\end{align}

\subsection{Evaluation Beyond Conventional Loss Metrics}
\label{sec:evaluation-methods}
Previous studies in the evaluation of reasoning models employ primarily two approaches: LLM-as-a-Judge for qualitative assessment \cite{zheng_judging_2023} and standard accuracy benchmarks (e.g., GSM8K, MATH) for quantitative validation. Loss-based metrics capture how well the model predicts tokens with respect to training data and are useful proxies for optimization progress; however, they fail to fully capture the reasoning quality and interpretability of generated responses. In particular, improvements in loss may not directly correspond to improvements in reasoning structure, factual reliability, or task completion.

To address these limitations, we adopt a complementary evaluation paradigm that leverages large language models (LLMs), such as GPT-4.1, as external judges. The evaluation criteria are scored on a scale from 0 to 10. More details regarding the specific prompt used can be found in Appendix~\ref{sec:GPT_judge_prompt_inst}.

\begin{itemize}
    \item \textbf{Coherence:} Measures the logical consistency and structural soundness of the reasoning process: logical flow between steps, structural organization, and consistency of assumptions/symbols.

    \item \textbf{Accuracy:} Evaluates the degree to which both intermediate reasoning and the final solution align with the correct answer, including factual correctness, domain knowledge application, reasoning validity, and final answer correctness.

    \item \textbf{Goal Completion:} Assesses whether the model employs valid reasoning strategies toward solving the problem, including strategy usefulness, progress toward solution, and partial success recognition.
\end{itemize}

\section{Experiments}

\subsection{Setup and Baselines}
We conducted experiments to evaluate the effectiveness of HiPO on math reasoning tasks, as these have been a strong contender in the research community for evaluating reasoning capabilities of language models. All experiments are performed on Qwen-2.5-7B Instruct and Llama-3.1-8B-Instruct base models, although our framework is designed to be model-agnostic.

\paragraph{Training Details:}
We implement HiPO, which consists of segment-level losses for each segment and preference loss for the entire response. Each output is decomposed into three segments---refined query ($Rq$), metacognitive thinking ($Mt$), and answer ($A$). Log-probabilities are computed separately for each segment as well as for the full response. Training is performed using AdamW with a learning rate of $1\times10^{-5}$ and sequence length of 512 tokens. Reference models are kept frozen, and the policy model is optimized end-to-end.

\paragraph{Weight Configurations:}
We formalize training regimes in terms of a $m \times n$ configuration matrix $C$, where each row corresponds to one training instance and each column specifies a hyperparameter. We define $n$ = \{$Rq\_weight$, $Mt\_weight$, $A\_weight$, $y\_weight$, $lr$, $epochs$\}, capturing both segment-level supervision weights and core optimization settings.

Within this space, we explore two regimes:
\begin{itemize}
    \item \textbf{Stepwise training.} The same model is trained sequentially across multiple rows of the configuration matrix $C$, each specifying a different weighting scheme and optimization schedule. We use the order ($Rq \rightarrow Mt \rightarrow Rq+Mt$).
    \begin{table}[H]
        \centering
        \resizebox{0.95\columnwidth}{!}{%
        \begin{tabular}{lcccccc}
        \toprule
        Regime & wt.$Rq$ & wt.$Mt$ & wt.$A$ & $y$ & lr & epochs \\
        \midrule
        Rq-bias & 0.60 & 0.15 & 0.15 & 0.10 & $1\mathrm{e}{-5}$ & 5 \\
        Mt-bias & 0.20 & 0.50 & 0.20 & 0.10 & $8\mathrm{e}{-6}$ & 5 \\
        Rq+Mt-bias & 0.35 & 0.30 & 0.15 & 0.25 & $5\mathrm{e}{-6}$ & 5 \\
        \bottomrule
        \end{tabular}%
        }
        \caption{Stepwise Training Weight Configuration.}
        \label{tab:configuration_matrix}
    \end{table}

    \item \textbf{Individual training.} The configuration matrix contains a single nonzero segment weight, and the model is trained under this fixed setting until convergence.
    \begin{table}[H]
        \centering
        \resizebox{0.75\columnwidth}{!}{%
        \begin{tabular}{lcccccc}
        \toprule
        Regime & $Rq$ & $Mt$ & $A$ & $y$ & lr & epochs \\
        \midrule
        Rq-only & 1.00 & 0.00 & 0.00 & 0.00 & $1\mathrm{e}{-6}$ & 5 \\
        Mt-only & 0.00 & 1.00 & 0.00 & 0.00 & $1\mathrm{e}{-6}$ & 5 \\
        A-only  & 0.00 & 0.00 & 1.00 & 0.00 & $1\mathrm{e}{-6}$ & 5 \\
        \bottomrule
        \end{tabular}%
        }
        \caption{Individual Training Weight Configuration.}
        \label{tab:individual_schedule}
    \end{table}
\end{itemize}

\paragraph{Baselines:}
\begin{enumerate}
    \item \textbf{Base}: The score of the base instruct model without any fine-tuning.
    \item \textbf{Standard DPO}: We calculate the DPO loss between the preferred and rejected answer (the version that does not consist of the broken-down segments).
    \item \textbf{Ablations}: Partial segment-level weighting and collapsing segments into a single preference score.
\end{enumerate}

\paragraph{Evaluation:}\label{para:eval}
For each weight configuration, we benchmark the model before and after training using standardized evaluation prompts. In addition to benchmark accuracy on GSM8K \cite{cobbe2021gsm8k}, Minerva \cite{NEURIPS2022_18abbeef}, AIME24\footnote{\url{https://huggingface.co/datasets/math-ai/aime24}}, Gaokao2023 \cite{zhang2024evaluatingperformancelargelanguage}, and MATH500 \cite{lightman2023letsverifystepstep}, we also adopt the complementary evaluation paradigm introduced in Section~\ref{sec:evaluation-methods}. These datasets span elementary to competition-level difficulty, allowing us to measure both easy-to-hard generalization and robustness of decomposition training.

\subsection{Results}

We first performed ablations to determine which components had the greatest effect on performance. All accuracies are reported using \textit{Final Answer Correctness}. Models were evaluated at temperature $0.1$ with sampling enabled (\texttt{do\_sample=True}).

\subsubsection{Models Considered}
We experiment with two model families: \textbf{Qwen2.5-7B-Instruct} and \textbf{Llama3.1-8B-Instruct}. Qwen was chosen because it is a competitive open-source model with strong chain-of-thought (CoT) reasoning abilities and prior exposure to mathematical datasets. Llama-8B provides a well-studied baseline for instruction-tuned models.

\subsubsection{Cross-Model Performance Analysis}

\paragraph{Individual Training Results}
The ablation studies (Appendix~\ref{sec:ablation-results}) across both model families demonstrate distinct optimization patterns for component-specific training. For Qwen2.5-7B-Instruct, the Rq-Only configuration achieves the maximum accuracy gain of +11.18\% on GSM8K with an average improvement of +4.46\% across all benchmarks. In contrast, Llama-3.1-8B-Instruct exhibits optimal performance with Mt-Only training, achieving a maximum accuracy gain of +11.00\% on AIME24 with an average of +1.57\% across benchmarks.

The A-Only configuration demonstrates consistently poor performance across both architectures. Qwen experiences the worst average accuracy change at -6.26\% with the lowest performance at -8.36\% on MATH500, while Llama shows comparable degradation with approximately -2.10\% average accuracy drop.

\paragraph{Stepwise Training Results}
The progressive training approach reveals complementary optimization strategies across model families. Qwen2.5-7B-Instruct achieves optimal performance through Mt-bias configuration, showing consistent accuracy improvements with a maximum gain of +9.18\% on GSM8K and an average of +3.68\% across all benchmarks (Table~\ref{tab:comparitive-study-stepwise-qwen}). Additionally, Rq+Mt bias allowed the model to gain a significant accuracy boost of +13.89\% on GSM8K, +6.4\% on MATH500, and +2.5\% on Gaokao with minimal accuracy drops across other benchmarks.

Llama-3.1-8B-Instruct exhibits different optimal configurations under stepwise training. The Rq-bias approach yields the strongest results with a maximum accuracy gain of +3.16\% and an average improvement of +1.83\% (Table~\ref{tab:comparitive-study-stepwise-llama}).

\begin{table*}[t]
    \centering
    \resizebox{\textwidth}{!}{%
        \begin{tabular}{lccccc}
            \toprule
                Config & Minerva & AIME24 & Gaokao2023 & GSM8K & MATH500 \\
            \midrule
            Base
              & 42.58\%
              & 7.33\%
              & 46.03\%
              & 81.80\%
              & 47.07\% \\

            DPO
              & 39.45\% [\textcolor{red}{-3.13}]
              & 5.66\% [\textcolor{red}{-1.66}]
              & 44.73\% [\textcolor{red}{-1.3}]
              & 81.34\% [\textcolor{red}{-0.45}]
              & 48.32\% [\textcolor{green}{+1.24}] \\

            HiPO-Rq-bias
              & 45.43\% [\textcolor{green}{+2.85}]
              & 9.00\% [\textcolor{green}{+1.67}]
              & 49.19\% [\textcolor{green}{+3.16}]
              & 83.27\% [\textcolor{green}{+1.47}]
              & 47.04\% [\textcolor{red}{-0.03}] \\

            HiPO-Mt-bias
              & 43.13\% [\textcolor{green}{+0.55}]
              & 0.00\% [\textcolor{red}{-7.33}]
              & 47.96\% [\textcolor{green}{+1.93}]
              & 83.55\% [\textcolor{green}{+1.75}]
              & 45.63\% [\textcolor{red}{-1.44}] \\

            HiPO-Rq+Mt-bias
              & 38.38\% [\textcolor{red}{-4.20}]
              & 0.66\% [\textcolor{red}{-6.67}]
              & 46.44\% [\textcolor{green}{+0.41}]
              & 85.52\% [\textcolor{green}{+3.72}]
              & 47.31\% [\textcolor{green}{+0.24}] \\
            \bottomrule
        \end{tabular}%
    }
    \caption{Stepwise Training results for Llama-3.1-8B-Instruct. Per-benchmark changes from base shown in brackets.}
    \label{tab:comparitive-study-stepwise-llama}
\end{table*}

\begin{table*}[t]
    \centering
    \resizebox{\textwidth}{!}{%
        \begin{tabular}{lccccc}
            \toprule
                Config & Minerva & AIME24 & Gaokao2023 & GSM8K & MATH500 \\
            \midrule
            Base
              & 48.26\%
              & 4.33\%
              & 58.61\%
              & 76.20\%
              & 60.07\% \\

            DPO
              & 48.81\% [\textcolor{green}{+0.55}]
              & 0.66\% [\textcolor{red}{-3.67}]
              & 57.53\% [\textcolor{red}{-1.08}]
              & 77.84\% [\textcolor{green}{+1.64}]
              & 60.92\% [\textcolor{green}{+0.85}] \\

            HiPO-Rq-bias
              & 47.66\% [\textcolor{red}{-0.60}]
              & 2.00\% [\textcolor{red}{-2.33}]
              & 59.49\% [\textcolor{green}{+0.88}]
              & 78.43\% [\textcolor{green}{+2.23}]
              & 59.60\% [\textcolor{red}{-0.47}] \\

            HiPO-Mt-bias
              & 51.02\% [\textcolor{green}{+2.76}]
              & 4.66\% [\textcolor{green}{+0.33}]
              & 62.18\% [\textcolor{green}{+3.57}]
              & 85.38\% [\textcolor{green}{+9.18}]
              & 62.62\% [\textcolor{green}{+2.55}] \\

            HiPO-Rq+Mt-bias
              & 47.65\% [\textcolor{red}{-0.61}]
              & 3.33\% [\textcolor{red}{-1.00}]
              & 61.11\% [\textcolor{green}{+2.50}]
              & 90.09\% [\textcolor{green}{+13.89}]
              & 66.50\% [\textcolor{green}{+6.43}] \\
            \bottomrule
        \end{tabular}%
    }
    \caption{Stepwise Training results for Qwen2.5-7B-Instruct. Per-benchmark changes from base shown in brackets.}
    \label{tab:comparitive-study-stepwise-qwen}
\end{table*}

\paragraph{Cross-Architecture Analysis}
The combined results reveal systematic patterns in hierarchical preference optimization effectiveness. Segment-specific training demonstrates clear advantages over traditional approaches, with Qwen achieving substantial improvements (average +4.46\% for Rq-only) and Llama showing more modest but consistent gains (average +1.57\% for Mt-only). The sequential training methodology successfully integrates multiple reasoning components, with Qwen achieving peak performance (+4.2\% average) through progressive Mt-focused training culminating in Rq+Mt-bias, while Llama demonstrates stability (+1.83\% average) through Rq-bias emphasis.

Task-specific performance patterns emerge across both models, revealing the nuanced benefits of hierarchical decomposition. Elementary reasoning tasks (GSM8K) show substantial improvements across both architectures, with Qwen reaching +13.89\% and Llama +3.16\% under optimal configurations. Competition-style problems (AIME24) demonstrate particular sensitivity to meta-thinking optimization, where Llama's Mt-only configuration achieves a +11.00\% improvement while Qwen's Mt-bias provides minor but consistent gains.

\section{Conclusion}

We presented HiPO (Hierarchical Preference Optimization), a novel framework that extends Direct Preference Optimization to structured reasoning tasks by decomposing responses into three semantically meaningful segments: Refined Query $Rq$, Meta-thinking $Mt$, and Answer $A$. Our approach enables targeted optimization of different reasoning aspects through segment-level auxiliary losses while maintaining DPO's computational efficiency and training stability. Experimental results show consistent and substantial gains over multiple benchmarks as well as improvements in coherence, accuracy, and goal completion. While the method shows promise for mathematical reasoning, results indicate sensitivity to dataset characteristics that future work should address. HiPO bridges stable preference optimization with hierarchical reasoning decomposition, opening new directions for training more interpretable and reliable reasoning-capable models.

While our experiments focus on mathematical reasoning, the hierarchical structure of HiPO naturally extends to retrieval-based systems. By separating reasoning components, HiPO provides a general framework that could enable targeted optimization of retrieval-aware reasoning in future work, bridging reasoning alignment and relevance optimization.

\bibliographystyle{plainnat}
\bibliography{references}


\newcommand{\drawdataset}[2]{%
  \foreach \i in {0,...,8} {
    \pgfmathparse{#1[\i]-\minrange+1}
    \pgfmathtruncatemacro{\angleindex}{\i+1}
    \coordinate (d#2\i) at (\angleindex * 360/\naxes:\pgfmathresult);
  }
  \draw[fill=#2, opacity=.1]
    (d#20) \foreach \i in {1,...,8} { -- (d#2\i) } -- cycle;
  \draw[thick,#2]
    (d#20) \foreach \i in {1,...,8} { -- (d#2\i) } -- cycle;
}
\newcommand{\radarchart}[5]{%
  \begin{tikzpicture}[scale=0.8]
    \def\naxes{9}
    \def\minrange{#1}
    \def\maxrange{#2}
    \def\dimensions{{Logical Flow},{Structural Organization},{Consistency},
                    {Domain Knowledge},{Reasoning Validity},
                    {Strategy Usefulness},{Progress Toward Solution},
                    {Partial Success},{Error Robustness}}
    \coordinate (origin) at (0,0);
    \pgfmathsetmacro{\difference}{\maxrange - \minrange + 1}
    \foreach \dim [count=\i from 1] in \dimensions {
      \pgfmathsetmacro{\angle}{\i * 360 / \naxes}
      \node at (\angle:\difference+0.5) {\scriptsize \dim};
      \draw[gray!60] (origin) -- (\angle:\difference);
    }
    \foreach \t in {1,...,\difference} {
      \draw[gray!30] (0,0) circle (\t);
      \pgfmathsetmacro{\lab}{\t+\minrange-1}
      \node[anchor=south] at (\t,0.2) {\scriptsize \lab};
    }
    #3
    #4
    \node[font=\bfseries] at (0,-\difference-1) {#5};
  \end{tikzpicture}
}

\appendix

\section{Additional Experimental Details}
\label{sec:ablation-results}

\def\QwenRaOnly{{8.413353115727004, 8.550445103857566, 8.550741839762612, 8.317507418397627, 7.8317507418397625, 8.53620178041543, 8.32492581602374, 8.001483679525222, 7.720771513353116}}
\def\QwenMtOnly{{8.589508002371073, 8.636929460580912, 8.640782454060462, 8.594250148192057, 8.219917012448132, 8.72554831061055, 8.57705986959099, 8.307646710136337, 8.08120924718435}}
\def\QwenRqOnly{{8.97836395969176, 8.997332542975697, 9.040901007705987, 8.938055720213397, 8.61962062833432, 9.024896265560166, 8.938944872554831, 8.683935981031416, 8.501066982809721}}
\def\QwenseqA{{8.82858837485172, 8.904804270462634, 8.920521945432977, 8.763048635824436, 8.378113879003559, 8.899169632265718, 8.767497034400948, 8.470047449584817, 8.249406880189799}}
\def\QwenseqB{{8.991691394658753, 9.048961424332344, 9.073293768545994, 8.955786350148369, 8.634421364985164, 9.056973293768547, 8.950445103857566, 8.700296735905045, 8.514243323442136}}
\def\QwenseqE{{8.929122182680901, 9.01067615658363, 9.0, 8.892348754448399, 8.511862396204034, 9.028173190984578, 8.88196915776987, 8.659549228944247, 8.429418742586002}}
\def\QwenDPO{{8.53206650831354, 8.564429928741093, 8.583432304038006, 8.505938242280285, 8.124109263657957, 8.65706650831354, 8.495546318289787, 8.199524940617577, 7.984263657957245}}
\def\QwenBase{{8.556346381969158, 8.58244365361803, 8.61832740213523, 8.54300118623962, 8.154507710557533, 8.690984578884935, 8.528173190984578, 8.243772241992882, 8.015124555160142}}

\def\LlamaRqOnly{{8.066765578635016, 8.061424332344213, 8.023145400593473, 7.8231454005934715, 7.413946587537092, 7.968842729970326, 7.898219584569733, 7.566765578635015, 7.262314540059347}}
\def\LlamaMtOnly{{8.113230035756853, 8.133492252681764, 8.089392133492252, 7.864719904648391, 7.465137067938022, 8.022348033373063, 7.962455303933254, 7.628426698450537, 7.337902264600715}}
\def\LlamaRaOnly{{8.035608308605342, 8.04213649851632, 8.018991097922848, 7.781305637982196, 7.391394658753709, 7.927299703264095, 7.87893175074184, 7.55459940652819, 7.240356083086054}}
\def\LlamaseqA{{8.313725490196079, 8.15032679738562, 8.235591206179441, 8.035353535353535, 7.663398692810458, 8.173202614379084, 8.160724896019014, 7.851158645276293, 7.4887106357694595}}
\def\LlamaseqB{{8.237358715050565, 8.054729327781082, 8.117192147531231, 7.9699583581201665, 7.57674003569304, 8.099345627602618, 8.055026769779893, 7.722784057108864, 7.361392028554432}}
\def\LlamaseqC{{8.272835112692764, 8.111209964412812, 8.118920521945434, 8.042408066429418, 7.640569395017794, 8.184460260972717, 8.126334519572953, 7.806346381969158, 7.476275207591933}}
\def\LlamaBase{{8.090071343638526, 8.10731272294887, 8.031807372175981, 7.831747919143877, 7.445303210463734, 7.99910820451843, 7.933412604042807, 7.667063020214031, 7.339179548156956}}
\def\LlamaDPO{{8.125222551928783, 8.128783382789317, 8.091988130563799, 7.911869436201781, 7.511572700296736, 8.065281899109792, 7.99080118694362, 7.6596439169139465, 7.378338278931751}}

We show the results of the GPT-4.1 evaluation (Section~\ref{sec:evaluation-methods}) for the models in various training regimes.

\begin{figure}[H]
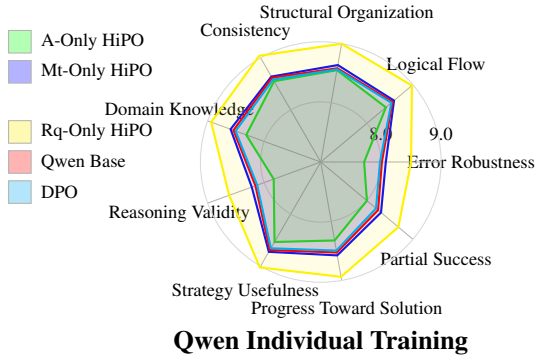

    \radarchart{8}{9}{%
      \drawdataset{\QwenRaOnly}{green}
      \drawdataset{\QwenMtOnly}{blue}
      \drawdataset{\QwenRqOnly}{yellow}
      \drawdataset{\QwenBase}{red}
      \drawdataset{\QwenDPO}{cyan}
    }{%
      \node[draw, fill=green, opacity=0.3, minimum size=0.3cm, label=right:\scriptsize A-Only HiPO] at (-5,2) {};
      \node[draw, fill=blue, opacity=0.3, minimum size=0.3cm, label=right:\scriptsize Mt-Only HiPO] at (-5,1.5) {};
      \node[draw, fill=yellow, opacity=0.3, minimum size=0.3cm, label=right:\scriptsize Rq-Only HiPO] at (-5,0.5) {};
      \node[draw, fill=red, opacity=0.3, minimum size=0.3cm, label=right:\scriptsize Qwen Base] at (-5,0) {};
      \node[draw, fill=cyan, opacity=0.3, minimum size=0.3cm, label=right:\scriptsize DPO] at (-5,-0.5) {};
    }{Qwen Individual Training}
    \caption{Rq-Only HiPO (yellow) consistently scores highest across all reasoning dimensions for Qwen, highlighting the impact of query refinement.}
    \label{fig:Qwen-ind-radar}
\end{figure}

\begin{figure}[H]
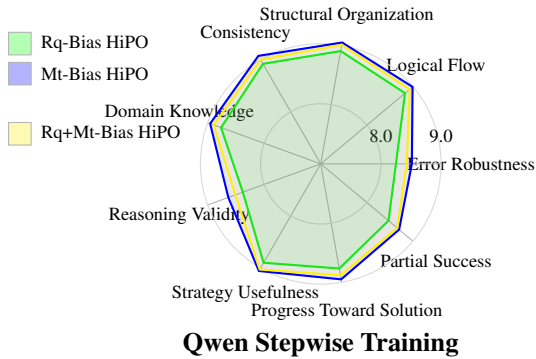

    \radarchart{8}{9}{%
      \drawdataset{\QwenseqA}{green}
      \drawdataset{\QwenseqB}{blue}
      \drawdataset{\QwenseqE}{yellow}
    }{%
      \node[draw, fill=green, opacity=0.3, minimum size=0.3cm, label=right:\scriptsize Rq-Bias HiPO] at (-5,2) {};
      \node[draw, fill=blue, opacity=0.3, minimum size=0.3cm, label=right:\scriptsize Mt-Bias HiPO] at (-5,1.5) {};
      \node[draw, fill=yellow, opacity=0.3, minimum size=0.3cm, label=right:\scriptsize Rq+Mt-Bias HiPO] at (-5,0.5) {};
    }{Qwen Stepwise Training}
    \caption{Rq+Mt-Bias HiPO (yellow) achieves the most balanced and consistently high scores for Qwen stepwise training.}
    \label{fig:Qwen-seq-radar}
\end{figure}

\begin{figure}[H]
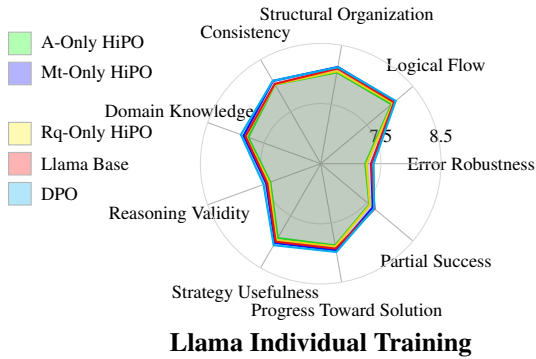

    \radarchart{7.5}{8.5}{%
      \drawdataset{\LlamaRaOnly}{green}
      \drawdataset{\LlamaMtOnly}{blue}
      \drawdataset{\LlamaRqOnly}{yellow}
      \drawdataset{\LlamaBase}{red}
      \drawdataset{\LlamaDPO}{cyan}
    }{%
      \node[draw, fill=green, opacity=0.3, minimum size=0.3cm, label=right:\scriptsize A-Only HiPO] at (-5,2) {};
      \node[draw, fill=blue, opacity=0.3, minimum size=0.3cm, label=right:\scriptsize Mt-Only HiPO] at (-5,1.5) {};
      \node[draw, fill=yellow, opacity=0.3, minimum size=0.3cm, label=right:\scriptsize Rq-Only HiPO] at (-5,0.5) {};
      \node[draw, fill=red, opacity=0.3, minimum size=0.3cm, label=right:\scriptsize Llama Base] at (-5,0) {};
      \node[draw, fill=cyan, opacity=0.3, minimum size=0.3cm, label=right:\scriptsize DPO] at (-5,-0.5) {};
    }{Llama Individual Training}
    \caption{Rq-Only HiPO (yellow) achieves the highest scores across most dimensions for Llama individual training.}
    \label{fig:llama-ind-radar}
\end{figure}

\begin{figure}[H]
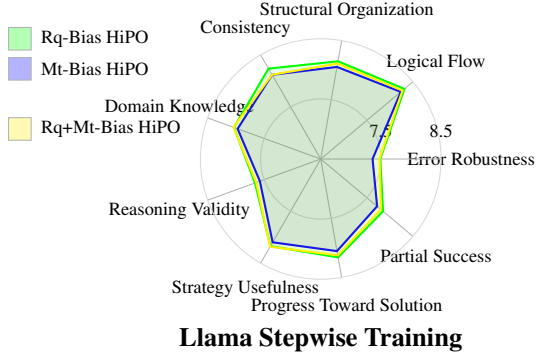

    \radarchart{7.5}{8.5}{%
      \drawdataset{\LlamaseqA}{green}
      \drawdataset{\LlamaseqB}{blue}
      \drawdataset{\LlamaseqC}{yellow}
    }{%
      \node[draw, fill=green, opacity=0.3, minimum size=0.3cm, label=right:\scriptsize Rq-Bias HiPO] at (-5,2) {};
      \node[draw, fill=blue, opacity=0.3, minimum size=0.3cm, label=right:\scriptsize Mt-Bias HiPO] at (-5,1.5) {};
      \node[draw, fill=yellow, opacity=0.3, minimum size=0.3cm, label=right:\scriptsize Rq+Mt-Bias HiPO] at (-5,0.5) {};
    }{Llama Stepwise Training}
    \caption{Rq+Mt-Bias HiPO (yellow) achieves the most balanced performance for Llama stepwise training.}
    \label{fig:llama-seq-radar}
\end{figure}

\section{Extended Ablation Results}

\begin{table}[H]
    \centering
    \resizebox{\columnwidth}{!}{%
    \begin{tabular}{lccccc}
        \toprule
            Config & Minerva & AIME24 & Gaokao2023 & GSM8K & MATH500 \\
        \midrule
        Base
          & 42.58\%
          & 7.33\%
          & 46.03\%
          & 81.80\%
          & 47.07\% \\

        DPO
          & 39.45\% [\textcolor{red}{-3.13}]
          & 5.66\% [\textcolor{red}{-1.66}]
          & 44.73\% [\textcolor{red}{-1.3}]
          & 81.34\% [\textcolor{red}{-0.45}]
          & 48.32\% [\textcolor{green}{+1.24}] \\

        Rq-only [\textcolor{green}{+4.46}]
          & 53.49\% [\textcolor{green}{+5.23}]
          & 1.33\% [\textcolor{red}{-3.00}]
          & 61.68\% [\textcolor{green}{+3.07}]
          & 87.38\% [\textcolor{green}{+11.18}]
          & 65.88\% [\textcolor{green}{+5.81}] \\

        Mt-only [\textcolor{green}{+0.98}]
          & 50.68\% [\textcolor{green}{+2.42}]
          & 3.33\% [\textcolor{red}{-1.00}]
          & 58.46\% [\textcolor{red}{-0.15}]
          & 80.00\% [\textcolor{green}{+3.80}]
          & 59.92\% [\textcolor{red}{-0.15}] \\

        A-only [\textcolor{red}{-6.26}]
          & 45.37\% [\textcolor{red}{-2.89}]
          & 0.00\% [\textcolor{red}{-4.33}]
          & 50.19\% [\textcolor{red}{-8.42}]
          & 67.90\% [\textcolor{red}{-8.30}]
          & 51.71\% [\textcolor{red}{-8.36}] \\
      \bottomrule
    \end{tabular}%
    }
    \caption{Individual Training ablation for Qwen2.5-7B-Instruct. Average change from base shown in brackets after config name.}
    \label{tab:comparitive-study-ind-qwen}
\end{table}

\begin{table}[H]
    \centering
    \resizebox{\columnwidth}{!}{%
    \begin{tabular}{lccccc}
        \toprule
            Config & Minerva & AIME24 & Gaokao2023 & GSM8K & MATH500 \\
        \midrule
        Base
          & 42.58\%
          & 7.33\%
          & 46.03\%
          & 81.80\%
          & 47.07\% \\

        DPO [\textcolor{red}{-1.06}]
          & 39.45\% [\textcolor{red}{-3.13}]
          & 5.66\% [\textcolor{red}{-1.66}]
          & 44.73\% [\textcolor{red}{-1.3}]
          & 81.34\% [\textcolor{red}{-0.45}]
          & 48.32\% [\textcolor{green}{+1.24}] \\

        Rq-only [\textcolor{red}{-1.96}]
          & 43.08\% [\textcolor{green}{+0.50}]
          & 0.00\% [\textcolor{red}{-7.33}]
          & 45.61\% [\textcolor{red}{-0.42}]
          & 78.44\% [\textcolor{red}{-3.36}]
          & 47.88\% [\textcolor{green}{+0.81}] \\

        Mt-only [\textcolor{green}{+1.57}]
          & 42.97\% [\textcolor{green}{+0.39}]
          & 18.33\% [\textcolor{green}{+11.00}]
          & 45.48\% [\textcolor{red}{-0.55}]
          & 79.92\% [\textcolor{red}{-1.88}]
          & 45.98\% [\textcolor{red}{-1.09}] \\

        A-only [\textcolor{red}{-2.10}]
          & 40.22\% [\textcolor{red}{-2.36}]
          & 3.33\% [\textcolor{red}{-4.00}]
          & 44.29\% [\textcolor{red}{-1.74}]
          & 80.14\% [\textcolor{red}{-1.66}]
          & 46.34\% [\textcolor{red}{-0.73}] \\
      \bottomrule
    \end{tabular}%
    }
    \caption{Individual Training ablation for Llama-3.1-8B-Instruct.}
    \label{tab:comparitive-study-ind-llama}
\end{table}

\section{Hyperparameters and Training Settings}

\begin{table}[H]
    \centering
    \begin{tabular}{lcccccc}
    \toprule
    Config & Wt.$Rq$ & Wt.$Mt$ & Wt.$A$ & Wt.$Y$ & lr & epochs\\
    \midrule
    Rq-Only   & 1.00 & 0.00 & 0.00 & 0.00 & $1\mathrm{e}{-6}$ & 5 \\
    Mt-Only   & 0.00 & 1.00 & 0.00 & 0.00 & $1\mathrm{e}{-6}$ & 5 \\
    A-Only    & 0.00 & 0.00 & 1.00 & 0.00 & $1\mathrm{e}{-6}$ & 5 \\
    Rq-bias   & 0.60 & 0.15 & 0.15 & 0.10 & $1\mathrm{e}{-6}$ & 5 \\
    Mt-bias   & 0.20 & 0.50 & 0.20 & 0.10 & $1\mathrm{e}{-6}$ & 5 \\
    Rq+Mt-bias & 0.35 & 0.30 & 0.15 & 0.25 & $1\mathrm{e}{-6}$ & 5 \\
    \bottomrule
    \end{tabular}
    \caption{Individual Training Weight Configurations.}
    \label{tab:individual-schedule-weights}
\end{table}

\begin{table}[H]
    \centering
    \begin{tabular}{clcccccc}
    \toprule
    Step & Config & Wt.$Rq$ & Wt.$Mt$ & Wt.$A$ & Wt.$Y$ & lr & epochs\\
    \midrule
    0 & Rq-bias    & 0.60 & 0.15 & 0.15 & 0.10 & $1\mathrm{e}{-5}$ & 5 \\
    1 & Mt-bias    & 0.20 & 0.50 & 0.20 & 0.10 & $8\mathrm{e}{-6}$ & 5 \\
    2 & Rq+Mt-bias & 0.35 & 0.30 & 0.15 & 0.25 & $5\mathrm{e}{-6}$ & 5 \\
    \bottomrule
    \end{tabular}
    \caption{Stepwise Training Weight Configuration.}
    \label{tab:sequential-schedule-weights}
\end{table}

\section{GPT Dataset Prompt Instruction}
\label{sec:GPT_dataset_prompt_inst}

\begin{tcolorbox}[title=Dataset Generation Prompt (1/2), sharp corners, colback=gray!5]
You are given a dataset of instructions, two model outputs (\texttt{output\_a} and \texttt{output\_b}).

Your task is to rewrite both outputs into the following cognitive structure:

1. Refined Query ($R_q$) --- Rewrite the original query into an elaborate one that contains more explanations or context for answering the original query.

2. Meta-Thinking ($M_t$) --- Provide structured reasoning steps that logically lead to the answer.

3. Refined Answer ($A$) --- Give the final, polished response that directly addresses the query, based on $M_t$.
\end{tcolorbox}

\begin{tcolorbox}[title=Dataset Generation Prompt (2/2), sharp corners, colback=gray!5]
Format your response strictly as JSON with the following structure:

\texttt{\{"output\_a": \{"refined\_query": "...", "meta\_thinking": "...", "refined\_answer": "..."\},}

\texttt{"output\_b": \{"refined\_query": "...", "meta\_thinking": "...", "refined\_answer": "..."\}\}}

Maintain the preference relationship: if output\_a was originally preferred, ensure your rewritten output\_a remains of higher quality than output\_b. Do not add any text before or after the JSON.
\end{tcolorbox}

\section{GPT Judge Prompt Instruction}
\label{sec:GPT_judge_prompt_inst}

\begin{tcolorbox}[title=Judge Prompt, sharp corners, colback=gray!5]
You are an expert evaluator of mathematical reasoning. Given a problem and a model response, evaluate the response on the following criteria, each scored 0--10:

\textbf{Coherence:} Logical flow, structural organization, consistency.

\textbf{Accuracy:} Factual correctness, domain knowledge application, reasoning validity, final answer correctness.

\textbf{Goal Completion:} Strategy usefulness, progress toward solution, partial success recognition, error robustness.

Return a JSON object: \texttt{\{"coherence": \{...\}, "accuracy": \{...\}, "goal\_completion": \{...\}\}}
\end{tcolorbox}

\end{document}